\documentclass[10pt,journal,compsoc]{IEEEtran}

\usepackage{amsmath,amsfonts}
\usepackage{algorithm}
\usepackage{algorithmicx}
\usepackage{array}
\usepackage[caption=false,font=normalsize,labelfont=sf,textfont=sf]{subfig}
\usepackage{multirow}
\usepackage{booktabs}
\usepackage{textcomp}
\usepackage{stfloats}
\usepackage{url}
\usepackage{verbatim}
\usepackage{graphicx}
\usepackage{multirow}
\newcommand{\demph}[1]{\textbf{\textit{#1}}}
\usepackage[utf8]{inputenc} 
\usepackage[T1]{fontenc}    
\usepackage[breaklinks=true,colorlinks,bookmarks=false]{hyperref}
\usepackage{url}            
\usepackage{booktabs}       
\usepackage{amsfonts}       
\usepackage{nicefrac}       
\usepackage{microtype}      
\usepackage{amsthm}
\usepackage{amsmath}
\usepackage{graphicx}
\usepackage{amssymb}
\usepackage{diagbox}
\usepackage{multirow}
\usepackage{textcomp} 
\usepackage{pifont} 
\usepackage{bm}
\usepackage{color,xcolor}
\usepackage{caption}
\usepackage{wrapfig}
\usepackage{colortbl}
\usepackage{enumitem}
\usepackage{multirow}
\usepackage{algorithm, algpseudocode}
\usepackage{threeparttable}
\usepackage{mathrsfs}
\usepackage[misc]{ifsym}

\newcommand{\tabincell}[2]{\begin{tabular}{@{}#1@{}}#2\end{tabular}}

\usepackage{xcolor}

\definecolor{ForestGreen}{RGB}{34,139,34}
\definecolor{myyellow}{RGB}{181, 181, 27}

\ifCLASSOPTIONcompsoc
  \usepackage[nocompress]{cite}
\else
  \usepackage{cite}
\fi
\ifCLASSINFOpdf
\else
\fi
\def\method{ST-PAD}

\begin{document}
%

\title{Spatio-Temporal Fluid Dynamics Modeling via Physical-Awareness and Parameter \\ Diffusion Guidance for Governance}
%
%
%
%
\author{An Zhang,
        Sheng Chen,
        Pengcheng Wu,
        Chao Wu,
        Hao Wu,
        Fan Xu,
        and Kun Wang}
%
%


\markboth{IEEE TRANSACTIONS ON KNOWLEDGE AND DATA ENGINEERING}%
{Wang \MakeLowercase{\textit{et al.}}: Modeling Spatio-temporal Dynamical Systems with Neural Discrete Learning and Levels-of-Experts}
%



\IEEEtitleabstractindextext{%
\begin{abstract}

This paper proposes a two-stage framework named ST-PAD for physics-aware spatio-temporal continuum field modeling, aiming to achieve high-precision simulation and prediction of complex dynamical systems through spatio-temporal physical awareness and parameter diffusion guidance. The framework is designed to support the modeling of dynamic processes relevant to governance scenarios, including environmental monitoring, traffic management, and disaster prevention. In the upstream stage, we design a vector quantization reconstruction module with temporal evolution characteristics, ensuring balanced and resilient parameter distributions by introducing general physical constraints. In the downstream stage, a parameter-involved diffusion probabilistic network is utilized to generate high-quality future states of continuum fields, while enhancing the model's generalization ability by incorporating parameters from various physical settings. Extensive experiments on multiple benchmark datasets have verified the effectiveness and robustness of the ST-PAD framework, demonstrating its advantages over mainstream models in continuum field modeling and prediction, particularly in capturing local representations and maintaining strong out-of-distribution (OOD) generalization capabilities. These capabilities provide a potential computational foundation for understanding complex dynamic processes and supporting governance-related applications.
\end{abstract}

\begin{IEEEkeywords}
Spatio-temporal Data Mining, Fluid Dynamics, Out-of-distribution Generation.
\end{IEEEkeywords}}

\maketitle

\IEEEdisplaynontitleabstractindextext

%
\IEEEpeerreviewmaketitle

\ifCLASSOPTIONcompsoc
\IEEEraisesectionheading{\section{Introduction}\label{sec:introduction}}
\else
\section{Introduction}
\label{sec:introduction}
\fi

\IEEEPARstart{F}{luid} dynamics~\cite{li2020fourier,lu2021learning,wu2023solving} is a significant issue in earth science research. A deep understanding of fluid flows can greatly aid in explaining a vast array of earth phenomena, such as combustion dynamics \cite{candel2002combustion,kulsheimer2002combustion}, ocean dynamics \cite{xiong2023ai,wang2024xihe}, and molecular dynamics \cite{xu2024equivariant,hollingsworth2018molecular}, to name just a few. Typically, modeling fluid dynamics involves incorporating historical temporal information and entails the interaction between various spatial event states \cite{pfaff2020learning,lienen2022learning,shao2022transformer}, which can be further understood as a spatio-temporal modeling issue \cite{wu2023earthfarseer,wang2024modeling}.

\begin{figure}
  \centering
  \includegraphics[width=1\linewidth]{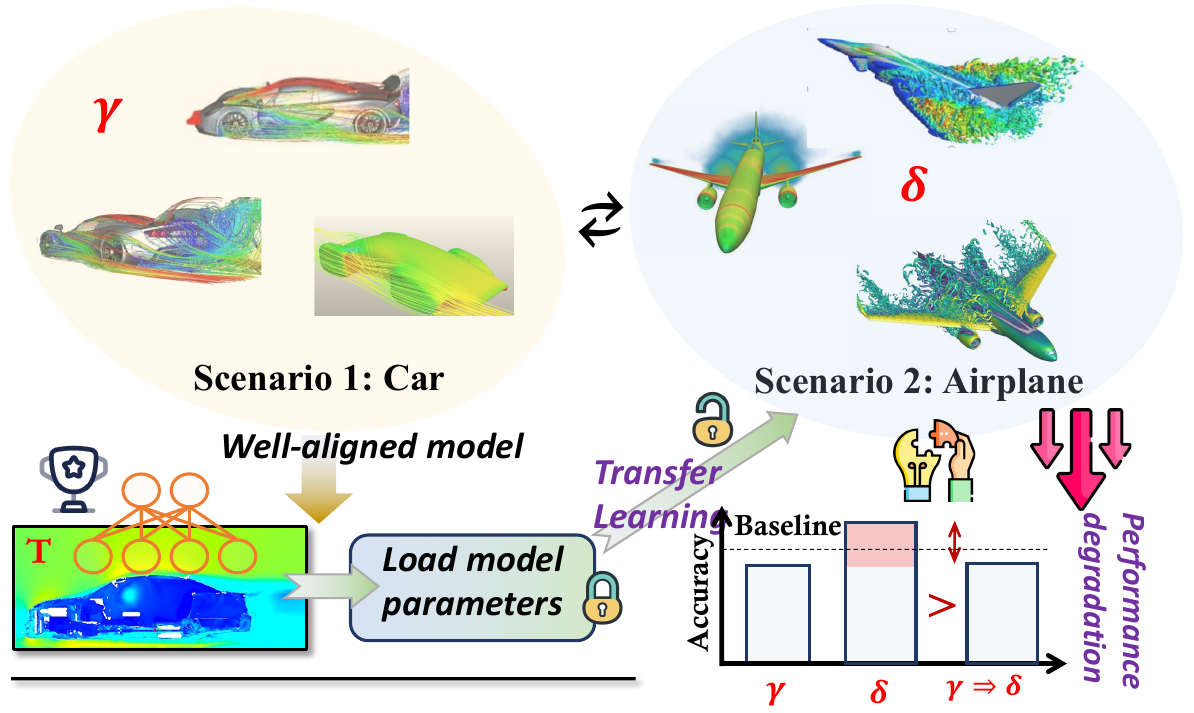}
       \caption{We observe that a model trained under scenario 1 does not perform well when directly transferred to scenario 2.}
  \label{fig:parametes}
\end{figure}

Modeling fluids presents a complex challenge due to their molecular structure, which lacks resistance to external shear forces. As a result, fluids are susceptible to deformation even under minimal forces, leading to highly nonlinear and chaotic behavior \cite{wu2023solving}. From a theoretical standpoint, the dynamics of fluids are governed by underlying partial differential equations (PDEs) \cite{long2018pde, chan2003variational}. Traditional approaches typically employ continuous Navier-Stokes equations \cite{constantin1988navier} to model fluid dynamics. Unfortunately, the high dimensionality of fluids introduces notorious inefficiency challenges to the solving process \cite{rabault2020deep,carlberg2013gnat}. Worse still, when modeling fluids, the training data encompasses both the dynamics themselves and the parameters of the dynamical system's PDEs. Training within a singular scenario struggles to generalize across systems with different parameter regimes \cite{takamoto2022cape,fotiadis2023disentangled,janny2024space}.

\begin{table*}[t]
\caption{Summary of existing high-performance numerical simulation models, PINNs, and pure data-driven models. The ST-PAD is distinguished by its unique ability to simultaneously offer the combined benefits of efficiency, PDE involvement, and generalizability.}
\vspace{-0.8em}
\label{table:intro}
\centering
\renewcommand\arraystretch{1.3}
\begin{tabular}{cc|cccccc}
\hline
\multicolumn{2}{c|}{\multirow{2}{*}{\begin{tabular}[c]{@{}c@{}}Traditional \& Physical-Informed NN\end{tabular}}} & \multicolumn{6}{c}{High-performances Methods}                                                                                                                                                                                                                                                                                                                                        \\ \cline{3-8} 
\multicolumn{2}{c|}{}                                                                                & \multicolumn{1}{c|}{FDS \cite{verda2021expanding}}                             & \multicolumn{1}{c|}{FLUENT \cite{rasinski2004creating}}                              & \multicolumn{1}{c|}{PINN \cite{karniadakis2021physics}}                                & \multicolumn{1}{c|}{MAD~\cite{huang2022meta}}                          & \multicolumn{1}{c|}{PPINN \cite{meng2020ppinn}}                                   & $\text{CAN-PINN}$ \cite{chiu2022can }                                              \\ \hline
\multicolumn{1}{c|}{\multirow{3}{*}{{Numerical calculation}
}}                    &    PDE-involve                         & \multicolumn{1}{c|}{\textcolor{green}{\checkmark} } & \multicolumn{1}{c|}{\textcolor{green}{\checkmark}}                      &\multicolumn{1}{c|}{}    & \multicolumn{1}{c|}{}   & \multicolumn{1}{c|}{}   &         \multicolumn{1}{c}{}                              \\ \cline{2-8} 
\multicolumn{1}{c|}{}                                         & Efficiency                               & \multicolumn{1}{c|}{ \textcolor{red}{\texttimes}}                                      & \multicolumn{1}{c|}{ \textcolor{red}{\texttimes}}  & \multicolumn{1}{c|}{}   & \multicolumn{1}{c|}{}             & \multicolumn{1}{c|}{}                         &    \multicolumn{1}{c}{}                                               \\ \cline{2-8} 
\multicolumn{1}{c|}{}                                         & Generalizability               & \multicolumn{1}{c|}{ \textcolor{red}{\texttimes}}          & \multicolumn{1}{c|}{ \textcolor{red}{\texttimes}}     & \multicolumn{1}{c|}{}                                        & \multicolumn{1}{c|}{}       & \multicolumn{1}{c|}{}        & \multicolumn{1}{c}{}      \\ \hline

\multicolumn{1}{c|}{\multirow{2}{*}{PINN}}               & PDE-involve                         & \multicolumn{1}{c|}{}                                         & \multicolumn{1}{c|}{}            & \multicolumn{1}{c|}{\textcolor{green}{\checkmark}}                                        & \multicolumn{1}{c|}{\textcolor{green}{\checkmark}}                                         & \multicolumn{1}{c|}{\textcolor{green}{\checkmark}}                                         &   \multicolumn{1}{c}{\textcolor{green}{\checkmark}}                                                  \\ \cline{2-8} 
\multicolumn{1}{c|}{}                                         & Generalizability                          & \multicolumn{1}{c|}{}            & \multicolumn{1}{c|}{}            & \multicolumn{1}{c|}{ \textcolor{red}{\texttimes}}  & \multicolumn{1}{c|}{ \textcolor{red}{\texttimes}}  & \multicolumn{1}{c|}{ \textcolor{red}{\texttimes}}       & \multicolumn{1}{c}{ \textcolor{red}{\texttimes}}                                                   \\ \hline \hline
\multicolumn{2}{c|}{\multirow{2}{*}{\begin{tabular}[c]{@{}c@{}}Data Driven\end{tabular}}} & \multicolumn{6}{c}{SOTA Methods}                                                                                                                                                                                                                                                                                                                                       \\ \cline{3-8} 
\multicolumn{2}{c|}{}                                                                                & \multicolumn{1}{c|}{TAU \cite{tan2023temporal}}                                      & \multicolumn{1}{c|}{PastNet \cite{wu2023pastnet}}                                      & \multicolumn{1}{c|}{Earthfarseer \cite{wu2023earthfarseer}}                                     & \multicolumn{1}{c|}{FNO \cite{li2020fourier}}                                      & \multicolumn{1}{c|}{DeepoNet \cite{lu2021learning}}                                      & ST-PAD (Ours)                                           \\ \hline
\multicolumn{1}{c|}{\multirow{3}{*}{{Purely Deep NN}}}                    & Efficiency                           & \multicolumn{1}{c|}{\textcolor{green}{\checkmark} } & \multicolumn{1}{c|}{\textcolor{green}{\checkmark} } & \multicolumn{1}{c|}{\textcolor{green}{\checkmark} }    & \multicolumn{1}{c|}{\textcolor{green}{\checkmark} } & \multicolumn{1}{c|}{\textcolor{green}{\checkmark} }                                        &    \multicolumn{1}{c}{\textcolor{green}{\checkmark} }                                                \\ \cline{2-8} 
\multicolumn{1}{c|}{}                                         & PDE-involve                               & \multicolumn{1}{c|}{ \textcolor{red}{\texttimes}} & \multicolumn{1}{c|}{ \textcolor{red}{\texttimes}}  &  \multicolumn{1}{c|}{ \textcolor{red}{\texttimes}} & \multicolumn{1}{c|}{\textcolor{green}{\checkmark} }                             &  \multicolumn{1}{c|}{ \textcolor{red}{\texttimes}}  &  \multicolumn{1}{c}{\textcolor{green}{\checkmark} }      \\ \cline{2-8} 
\multicolumn{1}{c|}{}                                         & Generalizability               &\multicolumn{1}{c|}{ \textcolor{red}{\texttimes}}   & \multicolumn{1}{c|}{ \textcolor{red}{\texttimes}}   & \multicolumn{1}{c|}{ \textcolor{red}{\texttimes}}   & \multicolumn{1}{c|}{ \textcolor{red}{\texttimes}}  & \multicolumn{1}{c|}{ \textcolor{red}{\texttimes}}     &    \multicolumn{1}{c}{\textcolor{green}{\checkmark} }         \\ \hline
\end{tabular}
\end{table*}

To this end, an emerging trend in deep fluid prediction involves training deep models to solve governing PDEs~\cite{lu2021learning, li2020fourier,krishnapriyan2021characterizing}. To address this, an emerging trend in deep fluid prediction focuses on training deep models to resolve governing partial differential equations (PDEs). A notable branch in this area is the advent of Physics-Informed Neural Networks (PINNs)~\cite{chen2022physics, raissi2019physics,yu2022gradient,xu2023physics}, which integrate deep learning principles with physics to tackle challenges in scientific computing, especially within the realm of fluid dynamics. PINNs enhance traditional neural network models by including a term in the loss function that accounts for the physical laws governing fluid dynamics, such as the Navier-Stokes equations~\cite{karniadakis2021physics,wang2023scientific,sitzmann2020implicit}. This inclusion guarantees that the network's predictions not only align with empirical data but also adhere to the fundamental principles of fluid dynamics. However, off-the-shelf PINNs often lack sufficient generalization capabilities, primarily due to customized loss function designs and the overlooking of network-specific parameter contexts \cite{takamoto2022cape,goring2024out}.

Although some studies attempt to learn fluid dynamics through data-driven approaches, these efforts \cite{wu2023pastnet, wu2023earthfarseer, xiong2023ai, li2020fourier, wu2023solving,wu2024spatiotemporal, lu2021learning, li2020fourier, li2023fourier, cao2021choose} predominantly employ a purely data-driven fashion for prediction, which does not permit generalization to PDE parameters in test data that differ from those in the training data. As illustrated in Fig \ref{fig:parametes}, we train fluid dynamics based on the motion state of cars under the parameter setting $\gamma$. We observe that it is challenging to transfer to the fluid motion of airplanes under the $\delta$ setting, denoted as $\gamma  \not\Rightarrow \delta$. Though few models consider PDE parameters, they are customized for specific neural networks \cite{yin2021leads,bereska2022continual}, and such efforts are susceptible to perturbing the stable features, which can lead to a loss of control over the fluid's evolutionary characteristics. The above dilemmas make it challenging for the deep fluid research community to design models with high generalization capabilities that cater not only to the initial conditions but also to different types of PDEs and their parameters. With this consideration, in our study, we explore \textit{whether it is possible to employ a universal fluid method that, while enhancing the network's generalization ability regarding PDE parameters, can also stably model the fluid dynamics change process, maintaining the model's stable understanding of the entire dynamics.}

To achieve this objective, we meticulously crafted a two-stage framework that incorporates spatio-temporal physical-awareness and parameter diffusion guidance, termed \textbf{ST-PAD}. Specifically, in the upstream, we design a Vector Quantized (VQ) reconstruction module with time-evolution characteristics. This module secures a reasonable parameter distribution by introducing a universal physical loss, such as thermal conduction \cite{lepri2003thermal} and convective diffusion loss \cite{gill1970exact}. To better capture the characteristics of the data, we incorporate Vector Quantized technology. Notably, since traditional VQ \cite{van2017neural,liu2021cross,fortuin2018som} techniques struggle with temporal tasks, we refine the entire VQ process with a Fourier module. This ensures observations can be made globally in the Fourier domain \cite{wu2023pastnet,guibas2021adaptive}, enabling the discovery of critical features throughout the temporal process, rather than allowing the model to focus on trivial parts.

Going beyond this process, we ensure the generation of high-quality image content while achieving parametric awareness of multiple physical settings through a parameter-involved diffusion probabilistic network \cite{bond2022unleashing,benny2022dynamic,choi2022perception,croitoru2023diffusion} in the downstream. Our approach provides a guiding framework for generating high-quality fluid dynamics parameter data \underline{for the first time}, without relying on complex mechanisms, \textit{i.e.,} channel attention mechanism \cite{takamoto2022cape}. The contributions of this paper can be summarized as follow:

\begin{itemize}[leftmargin=*]
    \item[\ding{182}] We propose a two-stage framework, ST-PAD, that systematically locks in the underlying task rules through ``pre-train'' in the upstream stage, and controls the model's generalization ability with downstream PDE parameter-involved diffusion ``fine-tuning''. To our knowledge, this is the first framework aimed at modeling the generalization capability for fluid dynamics.

    \item[\ding{183}] Our parameter-involved diffusion design, by integrating parameters into the diffusion process, enables the rapid generation of reliable data related to environmental settings, ensuring the capability for OOD generalization. This offers a dependable and quick sample generation solution for future fluid modeling. More importantly, as a plug-and-play and architecture-agnostic framework, this approach can be transferred to any Earth science task.

    \item[\ding{184}] Extensive experiments and iterative ablation studies demonstrate the strong generalization capabilities of ST-PAD. We compare its performance and generalizability against 12 of the current best models, and the results indicate that it can achieve superior performance by quickly adapting to different physical parameters.
\end{itemize}

\section{Related work} \label{related work}

\subsection{Spatiotemporal Predictive Learning}

In recent years, many novel architectures have emerged in the field of spatio-temporal (ST) predictive learning, various neural network architectures use different inductive biases to process data~\cite{oprea2020review}. They can be classified into three main categories: \textbf{(I) Convolutional Neural Network (CNN)-based architectures.} This research line focuses on extracting spatial features by employing CNN-based architecture~\cite{oh2015action, mathieu2016deep, chen2021deep, tulyakov2018mocogan}; \textbf{(II) Recurrent Neural Network (RNN)-based architectures}~\cite{srivastava2015unsupervised, babaeizadeh2018stochastic, seng2021spatiotemporal, wang2022predrnn}. Within this category, endeavors focus on optimizing the process of time-series data with some RNN and LSTM-based designs. \textbf{(III) Transformer-based architectures.} In a parallel vein, many models adopt Transformer-based frameworks to handle ST data~\cite{weissenborn2019scaling, kumar2020videoflow}. Going beyond these research lines, The hybrid model integrates the frameworks of multiple research lines, aiming to tap into the potential of each direction, while also inevitably introducing some additional parameters~\cite{wu2023pastnet, tan2023temporal}.

\subsection{Fluid Dynamics Modeling}
Early methods for fluid dynamics modeling primarily focused on utilizing tools like FDS \cite{verda2021expanding, wang2020applying}, FLUENT \cite{rasinski2004creating, baddeley1985components}, and Comsol \cite{pryor2009multiphysics, pepper2017finite} for solving PDEs. Due to the inefficiency of these methods and their difficulty in scaling to large datasets, many Physical Informed Neural Networks (PINNs) have subsequently attempted to combine physical loss and deep models to map historical data to future scenarios, which has become the mainstream paradigm in fluid modeling \cite{chen2022physics, raissi2019physics,yu2022gradient,xu2023physics}.

Though promising, PINNs heavily rely on domain-specific knowledge to introduce targeted PDE equations, making it challenging to enhance the generalization capability of fluid modeling. Then, many endeavors attempt to use purely data-driven deep models to model fluid dynamics. A resilient branch try to use neural operators to map historical data to future scenarios has become the mainstream paradigm in fluid modeling. This trend started with the introduction of DeepONet by Lu et al.~\cite{lu2021learning}, benefiting from the approximation theory of neural networks. Soon after, Li et al~\cite{li2020fourier}. proposed the FNO, which approximates the temporal integration process in fluid prediction by learning in the frequency domain. Building on this, Geo-FNO~\cite{li2023fourier} is specially optimized for irregular grids. The Galerkin Transformer~\cite{cao2021choose}, with enhanced Galerkin attention, exploits the excellent ability of attention mechanisms to handle long sequence data, calculating the correlation between features with linear complexity. GNOT~\cite{hao2023gnot} learns an efficient transformer by integrating grid information, equation parameters, and observation data. FactFormer~\cite{li2024scalable} accelerates the computation process by decomposing the attention mechanism into a low-rank form. NMO~\cite{anonymous2023neural} discovers finite subspaces from infinite-dimensional spaces through manifold learning, significantly improving the efficiency of PDE solutions and achieving state-of-the-art performance. However, the above method relies only on historical observation data, leading to poor generalization performance outside the distribution and lacks consideration of physical parameters (See Tab \ref{table:intro} for comparisons).

\subsection{Out-of-distribution Generalization}

The goal of Out-of-Distribution (OOD) generalization research line~\cite{volpi2018generalizing,mansilla2021domain,wu2022handling} is to improve the model's adaptability and robustness to new domains or data distributions not seen during training. This capability is especially critical for the practical implementation and deployment of applications, as it allows models to maintain high performance when facing unknown environments or tasks. OOD generalization is applied across multiple fields, including computer vision~\cite{li2022uncertainty}, natural language processing~\cite{gui2023joint}, and ST data mining~\cite{lu2024diversify}, helping models make better decisions and predictions in the face of the real world's complexity and variability. Many methods achieve OOD generalization through invariant learning~\cite{wang2023idea, wu2022discovering,wu2023discover}, aiming to construct invariant features across domains in latent space to minimize the misleading correlations caused by distribution changes. Moreover, techniques such as causal inference~\cite{wang2022out,wang2024nuwadynamics}, model selection~\cite{ye2021towards, yao2022improving}, and active learning~\cite{deng2023counterfactual, yang2024not} are used to enhance models' OOD generalization capabilities in real-world situations. We make our models more robust during the evolution of dynamical systems by combining ST physical awareness and guided parameter diffusion.

\section{Preliminary}
\noindent \textbf{Background of fluid dynamics.} In this article, we discuss partial differential equations (PDEs) in fluid dynamics defined over a time dimension $t \in [0, T]$ and a spatial dimension $\mathbf{x} = [x, y] \in \mathcal{X} \subseteq \mathbb{R}^2$. The spatial dimension forms a $h \times w$ grid, with each point $(x, y)$ representing a specific location in space. This setup allows for numerical approximations of fluid dynamics equations and the representation of fluid behavior at different locations on the grid. The fluid dynamics equations are:

\begin{equation}
\begin{aligned}
    & \frac{\partial u}{\partial t} = F(t, x, y, u, \nabla u, \nabla^2 u, \ldots) \quad  \\
    & \text{s.t.} \quad (t, x, y) \in [0, T] \times \mathcal{X},
\end{aligned}
\end{equation}

\noindent where $u(t, x, y)$ represents the state of the fluid, which can be a function of physical quantities like velocity, pressure, or density. The left side of the equation, $\frac{\partial u}{\partial t}$, shows the rate of change of the fluid state over time. The function $F$ on the right side describes how the fluid state changes based on its current state and spatial variations ($u$ and its spatial derivatives $\nabla u, \nabla^2 u, \ldots$). This function can incorporate various physical processes, including the effects of fluid viscosity, pressure gradients, and external forces. The initial condition $u(0, x, y) = u_0(x, y)$ specifies the fluid state at each spatial location at $t = 0$, and the boundary condition $B[u](t, x, y) = 0$ sets rules for fluid behavior at the boundaries. These rules can include fixed values (Dirichlet boundary conditions) or fluid flow rates (Neumann boundary conditions)~\cite{srebric2008cfd,blocken2007cfd}.

\noindent \textbf{Problem definition:} ST-PAD addresses solving PDEs in fluid dynamics by mapping them into sequences of spatial-temporal fields, $\left\{u_k\right\}_{k=0}^N$. Each $u_k$ reflects the fluid's condition at time $t_k$ and coordinates $x, y$. With the timeline discretized into intervals $t = T/N$ and covering all $x, y$ combinations, we precisely depict fluid motion. Concretely, we aim to emulate numerical simulations through a function $M: \mathcal{X} \rightarrow \mathcal{Y}$, which transitions each state $u_k$ to $u_{k+1}$ on a $h \times w$ grid, using auto-regressive fashion, which predicts future states from a series of fluid field data tensors of length $\lambda$, generating a sequence $\left\{\bar{u}_k\right\}_{k=\lambda}^N$. Model performance is assessed by comparing these forecasts against ground-truth, using mean squared error during the training phase:

\vspace{-0.5em}
\begin{equation}
\mathcal{L}(\theta) = \sum_{k=\lambda}^N {\rm{MSE}}\;(u_k, \bar{u}_k),
\end{equation}
\vspace{-0.5em}

\noindent By optimizing model parameters $\theta$, we aim to reduce the difference between predictions and actual observations, improving prediction accuracy. This method considers both time autoregressive properties and spatial complexity, including all combinations of $x, y$ at $h \times w$ resolution, assuming $\lambda > 1$, enabling the model to start generating ST sequences with an initial input sequence, capturing fluid details over time in complex spaces.

\section{Methodology}
\begin{figure*}[t]
  \centering
  \includegraphics[width=1\linewidth]{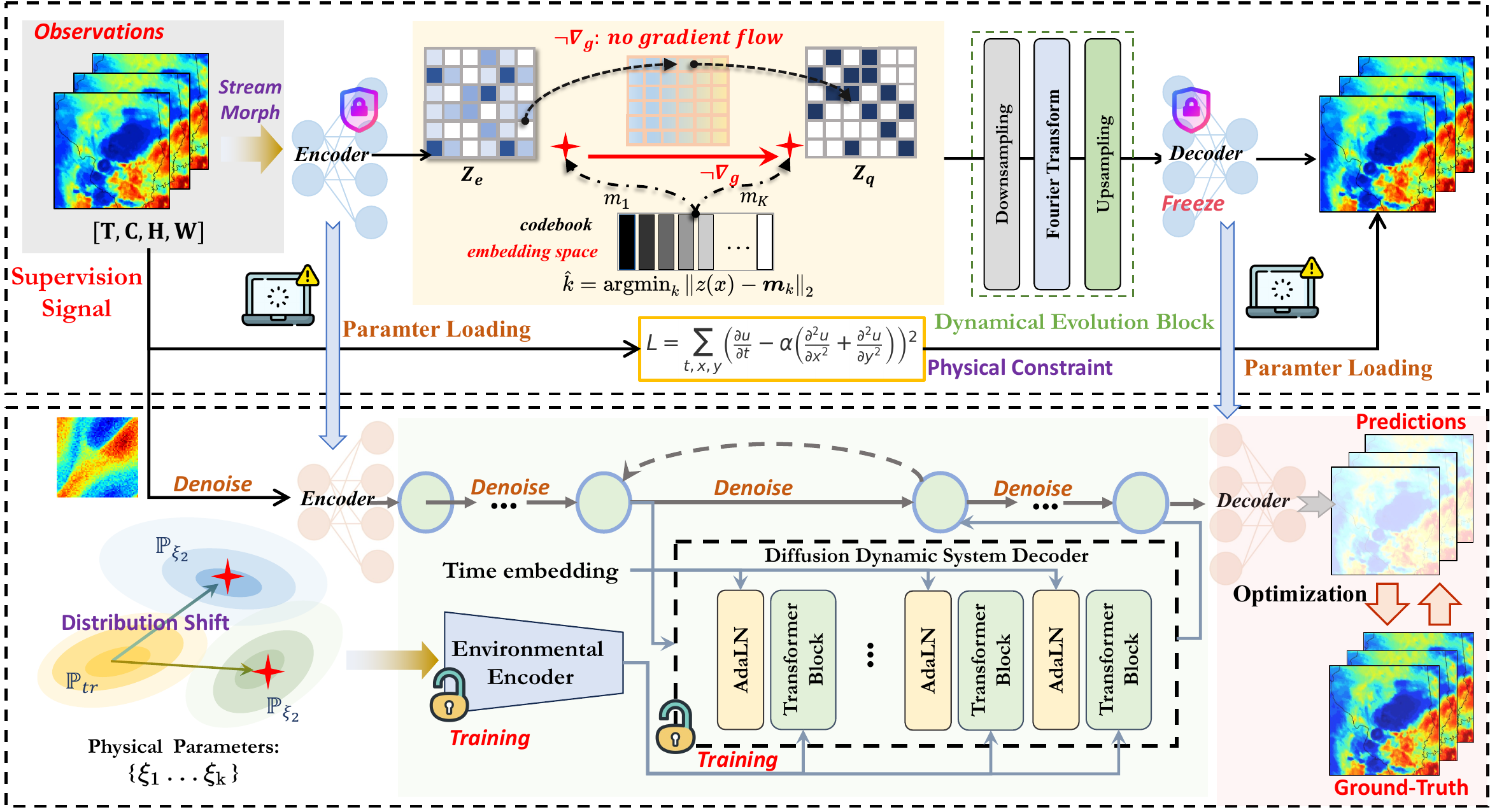}
  \vspace{-1.6em}
       \caption{Overview of the ST-PAD framework: The upper half demonstrates the upstream model, which locks in a reasonable range of model parameter distributions through self-supervision. The lower half of the model utilizes a parameter-involved diffusion model to enhance the model's generalization capabilities across different environmental settings.}
  \label{fig:moedl}
  \vspace{-1.4em}
\end{figure*}

In this section, we will systematically describe the technical details of the ST-PAD framework. The pipeline of our ST-PAD framework is presented in Fig. \ref{fig:moedl}, which takes historical fluid observations as input, that is, a spatio-temporal sequence. Specifically, we first present the entire process of upstream parameter locking in section \ref{sec4.1}. Subsequently, we detail the design of our parameter-involved diffusion in the downstream (Section \ref{sec4.2}). Finally, we demonstrate the theoretical guarantees of our model under causal theory (Section \ref{sec4.3}).

\subsection{Physical Constraint Pre-trained Model} \label{sec4.1}

Within the ST-PAD framework, we employs pre-training technique with physical laws to guarantee that the \underline{upstream} outputs are consistent with empirical data and adhere to established physical principles. This method improves generalization and prediction in varying or missing data scenarios by introducing a physics-driven loss function. In essence, it combines data-driven and physics-driven learning for accurate predictions in complex environments. Unlike PINNs, which heavily relies on explicit physical equations, ST-PAD capture physical correlations with only partially known equations (Showcased later in Section~\ref{sec:main}). Specifically, we utilize a self-supervised reconstruction mechanism guided by physical constraints to pinpoint a set of parameter weights that are relatively reasonable. This approach serves to initialize the parameters within a "better" attraction basin, positioning them near a local minimum that is conducive to enhanced generalization \cite{you2021graph, you2020graph}. We then reuse the encoder and decoder parameters for downstream tasks:
 
\noindent\textbf{Encoder Block.} The encoder consists of $L_e$ ConvNormReLU blocks, engineered for efficient spatial signal acquisition. Initiated with $\mathcal{Z}^{(0)} = \mathcal{X}$ and utilizing an activation function $\sigma$. Mathematically, the transformation for each stage $i$, where $1 \leq i \leq L_e$, is formalized as:
\begin{equation} 
\begin{gathered}
\mathcal{Z}^{(i)} = \sigma\left( \operatorname{GroupNorm}\left( \operatorname{ReLU}\left( \operatorname{Conv2d}\left( \mathcal{Z}^{(i-1)} \right) \right) \right) \right), \\
\end{gathered}
\end{equation}
\noindent where $\sigma$ represents the activation function, and the $\operatorname{Conv2d}$, $\operatorname{ReLU}$, and $\operatorname{GroupNorm}$ operations denote 2-dimensional convolution, rectified linear unit activation, and group normalization, respectively, and $\mathcal{Z}^{(i-1)}$ and $\mathcal{Z}^{(i)}$ denote the input and output of the $i$-th block with the shapes $(t, c, h, w)$ and $(t, \hat{c}, \hat{h}, \hat{w})$, respectively.

\noindent\textbf{Vector Quantization Discrete Encoding Module.} To optimize the embedding space and reduce computational costs while maintaining the accuracy of encoder output features, we use a vector quantization (VQ) discretization module. Noting that the output of encoder is $\mathcal{Z}^{e}$, then, it shape transforms from the form $(t, \hat{c}, \hat{h}, \hat{w})$ to $(L, D)$, where $L = h \times w$ and $D = t \times c$. In this process, each embedding vector \(\bm{z}\) from $\mathcal{Z}^{e}$ is meticulously mapped to the closest point in the memory bank, optimizing for precision in the embedding space. The memory bank is empirically structured to contain \(K\) embeddings, denoted as $\bm{\mathcal{M}}^{(K_e)}$ = \(\{\bm{m}_1, \cdots, \bm{m}_{K}\}\). Consequently, we establish a Vector Quantization mapping, \(VQ\), to facilitate this process:

\begin{equation}
VQ(\bm{z})=\bm{m}_{\hat{k}}, \quad \text { where } \quad \hat{k}=\operatorname{argmin}_{k}\left\|\bm{z}(\bm{x})-\bm{m}_{k}\right\|_{2},
\end{equation}

\noindent where each embedding $\bm{m}$ is concatenated to generate matrix $\bar{\mathcal{Z}}= VQ (\mathcal{Z}^{e})$. The mapping connects continuous vectors with given vectors in the memory bank to save the computational cost. Finally, the dimensions are reduced to $(t, \hat{c}, \hat{h}, \hat{w})$ as inputs to the dynamical evolution module.

\vspace{0.3em}
\noindent\textbf{Dynamical Evolution Block.} To capture the dynamic evolution of data and better understand data flows through complex transformations, we use dynamic evolution blocks. The dynamical evolution block consists of down-sampling, spectral domain processing and up-sampling three parts:
\begin{equation}
\begin{gathered}
\mathcal{Z}^{(i)} = \sigma\left( \phi\left( \text{Downsampling}\left(\mathcal{Y}^{(i-1)}, \theta \right) \right) \right), \\ \text{where} ; \mathcal{Y}^{(i-1)} = \int_{-\infty}^{+\infty} \mathcal{Z}^{(i-1)}(x) , dx, ; 1 \leq i \leq L_t
\end{gathered}
\end{equation}

\begin{equation}
\begin{gathered}
\mathcal{Z}^{(i)} = \sigma\left( \text{Fourier}\left( \text{Transform}\left(\mathcal{Z}^{(L_t)}, \theta' \right), \theta \right) \right)
\end{gathered}
\end{equation}

\begin{equation}
\begin{gathered}
\mathcal{Z}^{(i)} = \sigma\left( \text{Upsampling}\left( \Psi\left(\mathcal{Z}^{(i-1)}, \text{skip}^{(i)}, \lambda \right), \theta \right) \right), \\ \text{where} ; \Psi\left(a, b, \lambda\right) = \lambda a + (1-\lambda)b, ; L_t \geq i > 1
\end{gathered}
\end{equation}

\noindent where the feature representation \(\mathcal{Z}^{(i)}\) at each layer is obtained through a series of transformations applied to the features of the previous layer. These transformations include applying activation functions \(\sigma\) and \(\phi\) to increase the model's nonlinear capabilities. The transformation operation (\text{Transform}) processes data based on parameters \(\theta'\), while a specific fusion function \(\Psi\) combines the output of the previous layer with information from skip connections, adjusting their contributions through parameter \(\lambda\). The merging or fusion operation (\(\oplus\)) combines information from different sources to enrich the model's expressive capacity. Additionally, the introduced intermediate variables \(\mathcal{Y}^{(i-1)}\) and \(\mathcal{X}\) represent the transformed feature representations. Finally, \(\theta\) and \(\theta'\) are sets of parameters in different operations, adjusting the specific behaviors of their respective transformations.

\vspace{0.3em}
\noindent\textbf{Decoder Block.} To reconstruct the final output from features after dynamic evolution and restore the original form of the data, we use decoder blocks, our decoder contains $L_d$ unConvNormReLU blocks to output the final reconstructions $\mathcal{V}_{rec} = \mathcal{Z}^{(L_e + L_t +L_d)}$, where $L_{e}+L_{t}+1 \leq i \leq L_{e}+L_{t}+L_{d}$, In formulation, we can obtain:
\begin{equation}
\begin{aligned}
\mathcal{Z}^{(i)}= & \sigma\left(\operatorname{GroupNorm}\left(\operatorname{ReLU}\left(\operatorname{unConv2d}\left(\mathcal{Z}^{(i-1)}\right)\right)\right)\right),
\end{aligned}
\end{equation}

\noindent where $\mathcal{Z}_{rec}=\boldsymbol{\mathcal{Z}}_{t+1 : t + t'}=\left\{\mathcal{Z}_{t+1}, \cdots, \mathcal{Z}_{t+ t}'\right\}$, and the shape is ($t, c, h, w$).

\vspace{0.3em}
\noindent\textbf{Physical Constraint Loss Function.}To enhance the realism of model outputs and adherence to basic physical principles, we introduce a new method markedly different from traditional PINN. Unlike PINNs that rely on fully known physical equations, our approach works with partially known physical equations and emphasizes the application of Partial Differential Equations (PDE) by integrating physical constraint losses. This flexibility allows our model to make accurate predictions without complete understanding of the underlying physical processes, thus ensuring predictions dynamically conform to fundamental physical laws while broadening the application scope. The specific form of the loss function is:
\begin{equation}
L = \sum_{t,x,y} \left( \frac{\partial u}{\partial t} + u \cdot \nabla u - \alpha \nabla^2 u \right)^2,
\end{equation}
where \(u\) represents the velocity field and \(\alpha\) is a model parameter that represents the fluid's viscosity coefficient. This loss function constrains not only the temporal derivative of the predicted velocity field \( \frac{\partial u}{\partial t} \) but also includes constraints on the convection term \( u \cdot \nabla u \) and the diffusion term \( \alpha \nabla^2 u \). By minimizing this loss function, we can guide the model to learn dynamic fields that evolve within the framework of physical laws.Unlike traditional PINNs that require fully known PDEs, our upstream pretraining only assumes partial knowledge of physical processes. For example, in TaxiBJ+, we apply continuity constraints without requiring exact traffic flow equations. This design enables the model to maintain physical plausibility even when full governing equations are unavailable or inapplicable, especially in real-world systems where PDEs are only partially understood or data-driven approximations are needed.

\subsection{Parameter-involved Diffusion for OOD Generation}\label{sec4.2}

Intuitively, the concept of parameter-involved diffusion can be viewed as a generative process that integrates environmental parameters (e.g., viscosity, temperature) directly into the noise-to-sample decoding process. Instead of predicting future states from scratch, the model learns to denoise from a noisy prior distribution conditioned on these physical parameters, enabling it to generate fluid behaviors aligned with different physical settings. This formulation allows the model to generalize across systems with varying parameter regimes without retraining.

In downstream tasks, we reuse the encoder and decoder parameters trained in the upstream stage. This strategy significantly saves resources needed for training and greatly improves the model's learning efficiency on specific downstream tasks. Thanks to this parameter reuse, the model can transfer knowledge more efficiently between different tasks. This section introduces a novel parameter participation diffusion model design within the ST-PAD framework. It aims to enhance the model's generalization ability in different environmental settings, especially in OOD scenarios. Through an environment encoder and time embedding mechanism, the model captures environmental conditions (such as temperature and concentration) affecting fluid behavior and the dynamics of fluid changes over time. Moreover, the Diffusion Dynamic System Decoder (DDSD) predicts future dynamics of fluid states by combining environmental features and time-embedded data. The model trains by minimizing a comprehensive loss function to ensure physical consistency in predicted states and adherence to fundamental fluid dynamics laws, achieving accurate and reliable fluid dynamic simulation and prediction in complex environmental scenarios. Finally, we detail the component specifics in turn.

\noindent\textbf{Environmental Encoder and Time Embedding.} The environmental encoder extracts environmental feature vectors \(\mathcal{E}\) from the input observation data sequence \(\mathcal{X}\), capturing the environmental conditions that affect fluid behavior, such as temperature and concentration, The time embedding module encodes time information \(\tau\) into a feature vector \(\mathcal{T}\), to capture the dynamics of fluid changes over time. The mathematical definition is shown below:
\begin{equation}
\mathcal{E} = \text{EnvEncoder}(\mathcal{X}), \mathcal{T} = \text{TimeEmbedding}(\tau),
\end{equation}
where \(\mathcal{E}\) represents the environmental features encoded from the observation data \(\mathcal{X}\).
\noindent\textbf{Diffusion Dynamic System Decoder (DDSD).} The DDSD predicts future fluid states' dynamics by combining environmental features \(\mathcal{E}\) and time embeddings \(\mathcal{T}\). This module simulates a conditional reverse Markov chain to denoise and restore fluid states:
\begin{equation}
\mathcal{Z}_{\theta}^{(t)} = f{\theta}\left(\mathcal{Z}^{(t+1)}, \mathcal{E}, \mathcal{T}_t\right) + \epsilon^{(t)} \sqrt{1-\beta_t}, \quad t=T,\ldots,1,
\end{equation}
where \(f_{\theta}\) is a parameterized neural network function, \(\epsilon^{(t)}\) is the noise term, and \(\beta_t\) is the noise scheduling parameter at time step \(t\). The training objective is achieved by minimizing the following loss function:
\begin{equation}
\begin{aligned}
\mathcal{L}_{\text{DDSD}}(\theta) &= \mathbb{E}_{q(\mathcal{Z}_{0:T}|\mathcal{X})}\Bigg[\log p_{\theta}(\mathcal{V}|\mathcal{Z}_0) \\
&\quad - \sum_{t=1}^{T} \text{KL}\big(q(\mathcal{Z}_{t-1}|\mathcal{Z}_t, \mathcal{X}) \,||\, p_{\theta}(\mathcal{Z}_{t-1}|\mathcal{Z}_t)\big)\Bigg],
\end{aligned}
\end{equation}
To reinforce physical consistency and ensure the predicted states follow the laws of fluid dynamics, a physical law regularization term \(\mathcal{L}_{\text{phys}}\) is further integrated:
\begin{equation}
\mathcal{L}{\text{phys}}(\mathcal{Z}, \mathcal{P}) = \left|\nabla \cdot (\mathcal{Z} \otimes \mathcal{P}) - \nabla \cdot (\mathcal{Z}{\text{pred}} \otimes \mathcal{P}_{\text{pred}})\right|_2^2,
\end{equation}
Here, \(\mathcal{Z}\) and \(\mathcal{Z}_{\text{pred}}\) represent the velocity fields of the real and predicted states, respectively, while \(\mathcal{P}\) and \(\mathcal{P}_{\text{pred}}\) correspond to their respective pressure fields.

\subsection{Theoretical Guarantee}\label{sec4.3}

In this section, we examine and assess the rationality of the model from a causal perspective. Intuitively, the downstream diffusion process can be viewed as an introduction of environmental variables into the diffusion process to enhance the model's generalizability. Specifically, our framework can be interpreted as the process of front-door adjustment \cite{wu2022discovering, li2022let, fang2024exgc, fang2024moltc}, as illustrated in the right half of Fig \ref{fig:motivation}:

\begin{figure}[h]
  \centering
  \includegraphics[width=1\linewidth]{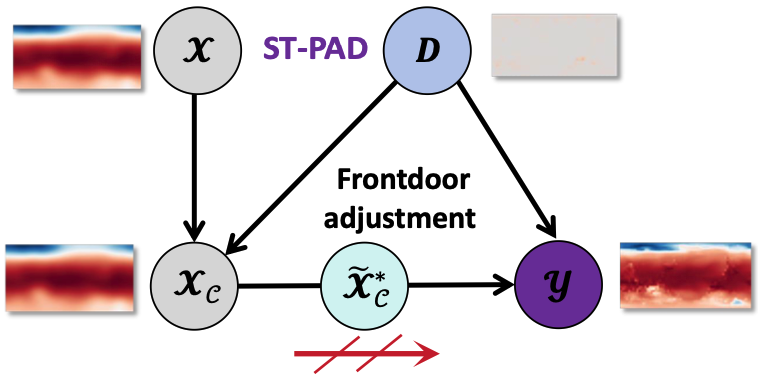}
       \caption{Diagram illustrating the front-door adjustment strategy in causal inference, with $\mathcal{X}$ representing the input, $\tilde {\mathcal X}_{\mathcal C}^*$, ${\mathcal X}_{\mathcal C}$ act as the surrogate variable of ${ {\mathcal X}_{\mathcal C}}$ and the causal part of $\mathcal{X}$, ${\mathcal D}$ denoting as a confounder.}
  \label{fig:motivation}
\end{figure}

\begin{itemize}[leftmargin=*]
    \item \(\mathcal{D}\) operates as a confounder, establishing an illusory correlation between \(\mathcal{X}_{\mathcal{C}}\) and \(\mathcal{Y}\). Here, \(\mathcal{X}_{\mathcal{C}}\) signifies the causal component incorporated within \(\mathcal{X}\).

     \item In the causal chain, \(\tilde{\mathcal{X}}_{\mathcal{C}}^*\) functions as the surrogate for the causal variable \(\mathcal{X}_{\mathcal{C}}\), enhancing it to fit the data distribution. Initially deriving from and encompassing \(\mathcal{X}_{\mathcal{C}}\), it represents the complete observations that should be apparent when observing the subpart \(\mathcal{X}_{\mathcal{C}}\). Additionally, \(\mathcal{X}_{\mathcal{C}}^*\) complies with the data distribution and maintains the essential knowledge of graph properties, thereby disconnecting any association between \(\mathcal{D}\) and \(\tilde{\mathcal{X}}_{\mathcal{C}}^*\). Hence, \(\tilde{\mathcal{X}}_{\mathcal{C}}^*\) aptly serves as the intermediary, which in turn influences the model's predictions for $\mathcal{Y}$.

\end{itemize}

\noindent In our front-door adjustment framework, we apply \textbf{do-calculus} to the variable \({\mathcal{X}}_{\mathcal{C}}\) to counteract the spurious correlations prompted by \({\mathcal{D}} \to {\mathcal{Y}}\). This elimination is effected by summing over the potential surrogate observations \(\tilde{X}_{\mathcal{C}}^*\). This strategy enables the connection of two distinguishable partial effects: \({\mathcal{X}}_{\mathcal{C}} \to \tilde{\mathcal{X}}_{\mathcal{C}}^*\) and \(\tilde{\mathcal{X}}_{\mathcal{C}}^* \to {\mathcal{Y}}\):

\noindent  

The process with \({\mathcal{X}}_{\mathcal{C}} \to \tilde{\mathcal{X}}_{\mathcal{C}}^*\) employs a parameter-involved diffusion approach to incorporate potential parameter settings. This methodology can be construed as a front-door adjustment designed to keenly capture and generate feasible scenario graphs based on physical parameter settings. \textit{By implementing the aforementioned strategy, we substantially augment the diffusion process's responsiveness to environmental variations, thus bolstering the model's generalization capacity.}

\subsection{Model Summary}
In summary, the ST-PAD framework employs a physical constraint pre-trained model and a physics-driven loss function to ensure model outputs align with empirical data and adhere to physical principles, thus enhancing generalization and prediction accuracy. Unlike methods fully relying on explicit physical equations, we use self-supervised learning and \textbf{\textit{partially known equations}} for modeling physical constraints. Moreover, the environmental encoder and time embedding mechanisms improve the model's generalization ability in various environmental settings, especially in OOD scenarios. The DDSD combines environmental features and time information to predict fluid dynamics, ensuring physical consistency in predictions through minimizing a comprehensive loss function, achieving accurate fluid dynamics simulations in complex environments. We summarize this as Algorithm~\ref{alg}.

\begin{algorithm}
\caption{ST-PAD Framework Process}
\label{alg}
\begin{algorithmic}[1]

\State \textbf{Input:} Spatio-temporal sequence $\mathcal{X}$

\State \textbf{Output:} Predicted dynamic fields $\mathcal{V}_{rec}$

\Comment{\textcolor{blue}{Upstream Parameter Locking}}
\State Initialize Encoder with ConvNormReLU blocks
\For{$i = 1$ to $L_e$}
    \State $\mathcal{Z}^{(i)} \gets$ Apply ConvNormReLU block on $\mathcal{Z}^{(i-1)}$
\EndFor
\State $\mathcal{Z}^{e} \gets$ Encoder's final layer output
\State Perform Vector Quantization on $\mathcal{Z}^{e}$

\State Initialize Dynamical Evolution Block
\For{$i = 1$ to $L_t$}
    \State Apply down-sampling, Fourier transform, up-sampling
\EndFor

\State Initialize Decoder
\For{$i = L_e + L_t + 1$ to $L_e + L_t + L_d$}
    \State $\mathcal{Z}^{(i)} \gets$ Apply unConvNormReLU block on $\mathcal{Z}^{(i-1)}$
\EndFor

\State Compute Physical Constraint Loss $L$

\Comment{\textcolor{blue}{Parameter-involved Diffusion}}
\State Extract environmental features $\mathcal{E}$ using EnvEncoder
\State Encode time information $\mathcal{T}$ using TimeEmbedding
\State Initialize DDSD
\For{$t = T$ to $1$}
    \State Predict $\mathcal{Z}_{\theta}^{(t)}$ using DDSD
\EndFor
\State Compute DDSD loss $\mathcal{L}_{\text{DDSD}}$
\State Compute Physical Law Regularization term $\mathcal{L}_{\text{phys}}$
\State \Return $\mathcal{Z}_{pred}$

\end{algorithmic}
\end{algorithm}

\section{Experiments}\label{sec:exp}
In this section, we showcase empirical evidence validating the efficacy of fluid methodologies, ST-PAD. The experimental investigations are designed to address the subsequent research inquiries:

\begin{itemize}
    \item \cal{RQ}1: How does the performance of the ST-PAD model compare to that of the current SOTA models?

    \item \cal{RQ}2: Is the ST-PAD model capable of achieving precise predictions structurally, as measured by metrics such as PSNR and SSIM?

    \item \cal{RQ}3: Can the design of our various components significantly enhance the capabilities of the ST-PAD model?

    \item \cal{RQ}4: Is the ST-PAD model capable of effectively ensuring the local fidelity of its predictions?
\end{itemize}

\subsection{Experimental Settings}

\noindent \textbf{Dataset.} We split the dataset into video, simulated and real-world datasets. The specific details are shown below. \textit{Video Data:} We chose KTH~\cite{schuldt2004recognizing} and TaxiBJ+~\cite{wu2023earthfarseer} as video datasets. \textit{Simulated Data:} We select representative Partial Differential Equation (PDE) datasets showcasing turbulence phenomena, including the Navier-Stokes (NS) equations~\cite{li2020fourier} and the Shallow-Water (SW) equations~\cite{takamoto2022pdebench}. Additionally, we choose iconic Computational Fluid Dynamics (CFD) datasets~\cite{wu2023spatio}, covering combustion dynamics (CD) datasets simulated with FDS and cylinder flow datasets. \textit{Real-world data:} We select the ERA5~\cite{bi2022pangu} and Weatherbench~\cite{rasp2023weatherbench} datasets to assess the model's capability in weather forecasting. Additionally, we choose the SEVIR~\cite{veillette2020sevir} dataset to analyze the model's ability to predict extreme events. Table~\ref{tab:dataset} presents the detailed statistical information of the dataset.

\begin{table}[h]
 \small
  \centering
  \def \arraystretch{1.2}
\setlength{\tabcolsep}{6pt}
        \caption{Dataset statistics. $N\_tr$ and $N\_te$ denote the number of instances in the training and test sets. The lengths of the input and prediction sequences are $I_l$ and $O_l$, respectively.}
    \begin{tabular}{l|cccccc}
    \hline
    \textbf{Dataset} & \textbf{$N\_tr$} & \textbf{$N\_te$} & \textbf{($C, H, W$)} & \textbf{$I_l$} & \textbf{$O_l$}  \\
    \hline
    TaxiBJ+ & 3555  & 445  & (2, 128, 128) & 12    & 12  \\
    KTH & 108717  & 4086  & (1, 128, 128) & 10    & 20  \\
    NS & 9000  & 1000  & (1, 64, 64) & 10    & 10  \\
    SW & 3555 & 445  & (2, 128, 128) & 12     & 12 \\
    CD   & 8000 & 1000  & (3, 128, 256) & 10    & 10  \\
    SEVIR & 4158  & 500   & (1, 384, 384) & 10    & 10  \\
    ERA5 & 6000  & 1500   & (1, 208, 333) & 10    & 10 \\
    WeatherBench & 2000  & 500   & (3, 128, 128) & 6    & 6 \\
    \hline
    \end{tabular}%
    \label{tab:dataset}
\end{table}

\noindent \textbf{Baselines.} To comprehensively assess the effectiveness of our proposed method, we select leading models in the fields of Physics-Informed Neural Network, Spatio-Temporal Forecasting and Neural Operators for comparison. For Physics-Informed Neural Network, we choose SOTA models as PINN~\cite{raissi2019physics}, MAD~\cite{huang2022meta}. For Spatio-Temporal Forecasting, we choose models such as ConvLSTM~\cite{shi2015convolutional}, SimVP~\cite{gao2022simvp}, Earthformer~\cite{gao2022earthformer}, TAU~\cite{tan2023temporal}, PastNet~\cite{wu2023pastnet}, and Earthfarseer~\cite{wu2023earthfarseer}. In the field of Neural Operators, we evaluate architectures like FNO~\cite{li2020fourier}, LSM~\cite{wu2023solving}, UNO~\cite{rahman2022u}, and F-FNO~\cite{tran2021factorized}. The details are as follows:

\begin{itemize}[leftmargin=*]

\item \textbf{PINN} is a deep learning model that embeds physical laws, such as conservation laws and dynamic equations, into the training process of neural networks to enhance their simulation and prediction capabilities for complex physical processes.

\item \textbf{MAD} is a novel approach for solving parametric partial differential equations (PDEs) that leverages meta-learning to efficiently adapt to various parameters, significantly enhancing solution accuracy and computational efficiency.

\item \textbf{ConvLSTM} is a neural network model for spatiotemporal sequence prediction, effectively capturing spatial and temporal dependencies in data by combining features of Convolutional Neural Networks (CNN) and Long Short-Term Memory networks (LSTM).

\item \textbf{SimVP} is a model designed for video prediction, effectively enhancing prediction accuracy and efficiency by simplifying the structured representation of the video prediction.

\item \textbf{Earthformer} is an exploratory approach that uses space-time Transformer models for Earth system forecasting, predicting dynamic changes in the Earth system by processing large-scale spatial-temporal data.

\item \textbf{TAU} is a model designed for video prediction, effectively enhancing prediction accuracy and efficiency by simplifying the structured representation of the video prediction.

\item \textbf{PastNet} is a spatiotemporal prediction model that integrates physical prior knowledge, designed to analyze and forecast spatiotemporal patterns in historical data for predicting future states.

\item \textbf{Earthfarseer} is specifically designed for spatiotemporal predictions in the field of Earth sciences, capable of processing large-scale geospatial data and predicting environmental and climate changes.

\item \textbf{FNO} is an operator learning framework that uses Fourier transforms to capture complex mappings between functions, aimed at solving partial differential equations and other continuous domain problems.

\item \textbf{LSM} is a method for learning and approximating spectral mappings, designed to enhance the efficiency and accuracy of learning operators by capturing the intrinsic spectral relationships between input and output data.

\item \textbf{UNO} is an operator learning model designed to generalize solutions for various partial differential equations, enhancing the model's generalization capability by learning representations of underlying physical rules.

\item \textbf{F-FNO} is an improved version of FNO that enhances computational efficiency through factorization techniques, aimed at efficiently and accurately solving complex continuous domain problems.

\end{itemize}

\begin{table*}[t] 
\caption{Model comparison with the state-of-the-arts over different evaluation metrics. We report the mean results from \textbf{five runs}. TaxiBJ+ metrics take MAE, all others are MSE metrics.}\label{tab:results1}
\vspace{-1.2em}
\setlength{\tabcolsep}{4pt}
\begin{center}
\def \arraystretch{1.2}
 \begin{tabular}{cccccccccccccc} 
 \toprule
    \multirow{2}{*}{\bf Backbone}  & \multicolumn{2}{c}{\bf Video Data} & \multicolumn{3}{c}{\bf Simulated Data}  &  \multicolumn{3}{c}{\bf Real-world data}   & \multirow{2}{*}{\bf Ranking} \\ 
    \cmidrule(l){2-3} \cmidrule(l){4-6} \cmidrule(l){7-9} 
    
     & \scriptsize \bf TaxiBJ+ (MAE)
     & \scriptsize \bf KTH (MSE) 
     
     & \scriptsize \bf SW (MSE)   
     & \scriptsize \bf NS (MSE)
     & \scriptsize \bf CD (MSE)

     & \scriptsize \bf WeatherBench (MSE) 
     & \scriptsize \bf ERA5 (MSE)
     & \scriptsize \bf SEVIR (MSE)    \\ 
    \midrule

    \multicolumn{9}{l}{{  \demph{ \it{\textcolor{blue}{Physics-Informed Neural Network}} }} }\\
    PINN~\cite{raissi2019physics}           & - & - & 0.6542 & 0.6857 & 2.8763 & - & - &-&13.00\\
    MAD~\cite{huang2022meta}                & - & - & 0.5834 & 0.4467 & 2.6532 & - & - &-&11.68\\

     \midrule
       \multicolumn{9}{l}{{  \demph{ \it{\textcolor{blue}{Spatio-temporal Predictive Models}} }} }\\
       
ConvLSTM~\cite{shi2015convolutional} & 5.5432 & 126.2322 & 0.3731 & 0.3334 & 0.9742 & 0.2983 & 4.1351 & 3.8124&7.88\\
SimVP~\cite{gao2022simvp} & 3.0212 & 40.9432 & 0.2121 & 0.1845 & 0.1932 & 0.1271 & 2.9371 & 3.4134 & 6.87\\
Earthformer~\cite{gao2022earthformer} & 3.1223 & 48.2643 & 0.1932 & 0.6543 & 0.2124 & 0.2124 & 3.5332 & 3.7159&8.63\\
TAU~\cite{tan2023temporal} & 2.9812 & 39.4312 & 0.2023 & 0.1723 & 0.1873 & 0.1345 & 2.9835 & 3.3397 & 6.50\\
PastNet~\cite{wu2023pastnet} & 2.7143 & 33.8433 & 0.2011 & 0.2144 & 0.1743 & 0.1312 & 3.0941 & 3.2322 & 6.50\\
Earthfarseer~\cite{wu2023earthfarseer} & 2.1109 & 31.8233 & 0.1921 & 0.1732 & 0.1384 & 0.098 & 2.8943 & 2.8314 & 4.37\\

 \midrule

           \multicolumn{9}{l}{{  \demph{ \it{\textcolor{blue}{Neural Operator Models}} }} }\\
FNO~\cite{li2020fourier}           & 5.4312 & 175.4432 & 0.1233 & 0.1546 & 1.5432 & 1.2432 & 2.8534 &4.4312&8.25\\
LSM~\cite{wu2023solving}           & 6.7532 & 174.3231 & 0.1543 & 0.1677 & 1.6873 & 1.0982 & 2.9763 &5.4983&9.38\\
UNO~\cite{rahman2022u}             & 3.2143 & 134.2131 & 0.1235 & 0.2012 & 1.4323 & 1.3321 & 2.5842 &3.0941&7.62\\
F-FNO~\cite{tran2021factorized}    & 5.3222 & 112.4312 & 0.1437 & 0.1924 & 1.8521 & 0.9874 & 8.9853 &3.9984&9.38\\
    \midrule
\textbf{ST-PAD} & \textbf{2.0729} & \textbf{29.9784} & \textbf{0.1129} & \textbf{0.1372} & \textbf{0.1231} & \textbf{0.0875} & \textbf{1.9872} & \textbf{2.7562} & \textbf{2.25}\\
\midrule
\vspace{-1.2em}
\end{tabular}
\end{center}
\end{table*}

\begin{figure*}[t]
  \centering
  \includegraphics[width=1\linewidth]{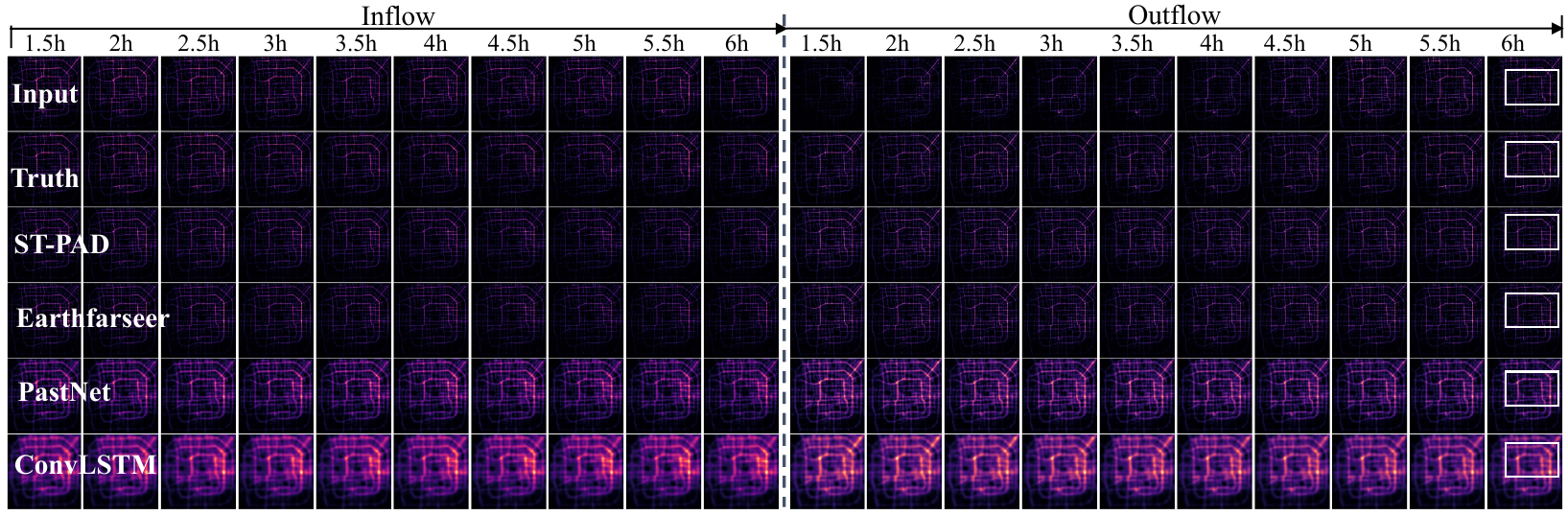}
  \vspace{-0.6em}
       \caption{Visualization on TaxiBJ+. For simplicity, we display the results of the last 10 frames.}
  \label{fig:main1}
  \vspace{-1.4em}
\end{figure*}

\noindent\textbf{Evaluation Metrics.} We adopt several metrics to assess the quality of our predictions, and each has its unique significance:
\begin{itemize}[leftmargin=*]
    \item \textbf{Mean Squared Error (MSE)}: This metric provides the average of the squares of the differences between the actual and predicted values. A lower MSE indicates a closer fit of the predictions to the true values. It's given by the equation:
    \begin{equation}
\text{MSE} = \frac{1}{N} \sum_{i=1}^{N} (V_{\text{true},i} - V_{\text{fut},i})^2
\end{equation}
where \( V_{\text{true},i} \) represents the true value, \( V_{\text{fut},i} \) denotes the predicted value, and \( N \) is the number of observations.

 \item \textbf{Mean Absolute Error (MAE)}: MAE provides the average of the absolute differences between the predicted and actual values. This metric is less sensitive to outliers than MSE, making it suitable for applications where extreme errors should not disproportionately affect the overall error measure.
\begin{equation}
\text{MAE} = \frac{1}{N} \sum_{i=1}^{N} |V_{\text{true},i} - V_{\text{fut},i}|
\end{equation}

 \item \textbf{Multi-Scale Structural Similarity (MS-SSIM)}: SSIM is designed to provide an assessment of the structural integrity and similarity between two images, \( x \) and \( y \). Higher SSIM values suggest that the structures of the two images being compared are more similar.
\begin{equation}
\text{SSIM}(x, y) = \frac{(2\mu_x\mu_y + c_1)(2\sigma_{xy} + c_2)}{(\mu_x^2 + \mu_y^2 + c_1)(\sigma_x^2 + \sigma_y^2 + c_2)}
\end{equation}
where \( \mu \) is the mean, \( \sigma \) represents variance, and \( c_1 \) and \( c_2 \) are constants to avoid instability.

 \item \textbf{Peak Signal-to-Noise Ratio (PSNR)}: PSNR gauges the quality of a reconstructed image compared to its original by measuring the ratio between the maximum possible power of the signal and the power of corrupting noise. A higher PSNR indicates a better reconstruction quality.
\begin{equation}
\text{PSNR} = 10 \times \log_{10} \left( \frac{\text{MAX}_I^2}{\text{MSE}} \right)
\end{equation}
where \( \text{MAX}_I \) is the maximum possible pixel value of the image.
\end{itemize}

\noindent \textbf{Implementation details.} In our experiment setup, we use the PyTorch framework and a 40GB A100 GPU for model training. We set the training period to 500 epochs, with an initial learning rate of 0.001. To control the training process more finely, we introduce a learning rate scheduler. It adjusts the learning rate by a decay rate of 0.5 every 100 epochs, adapting to different stages of model training. Also, we set our batch size to 10. On this basis, all our prediction models use an autoregressive prediction architecture.

\vspace{-0.6em}
\subsection{Main results}
\label{sec:main}
\vspace{-0.2em}

\begin{figure*}
  \centering
  \includegraphics[width=1\linewidth]{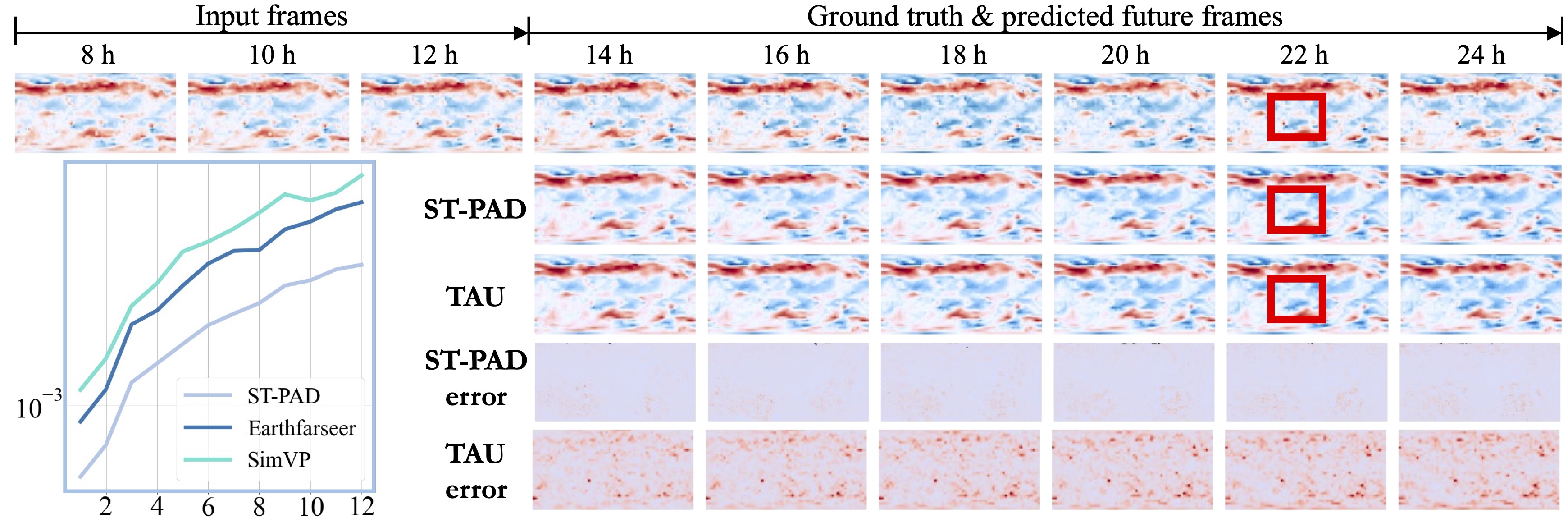}
       \caption{The figure shows input frames from 8 to 24 hours, along with their actual conditions and future frame predictions. The left side compares the performance of three models: ST-PAD, Earthforeseer, and SimVP. Red boxes highlight the local details that both models and the actual conditions successfully predicted. The bottom displays the error distribution of predictions made by the ST-PAD and TAU models.}
  \label{fig:main2}
\end{figure*}

We first focus on verifying the effectiveness of the ST-PAD. To systematically assess its capabilities, we select multiple datasets, and present the results in Table \ref{tab:results1}, Figure \ref{fig:main1} and Figure \ref{fig:main2}. From these results, we can list observations:

\textbf{Obs.\ding{182} The ST-PAD consistently outperforms the current mainstream models, including PINNs, spatio-temporal forecasting models, and neural operator models.} As demonstrated in Tab \ref{tab:results1}, ST-PAD maintains robust performance under the metrics MAE and MSE, achieving the top ranking across all frameworks. Specifically, on the video data set TaxiBJ+, ST-PAD attains a lower MAE (\underline{1.0494}) than the best model, Earthfarseer. On the CFD simulation data set, it achieves an MSE reduction of \underline{0.0153} compared to the best-performing Earthfarseer. Moreover, on the ERA5 weather data, it secures an MSE advantage of nearly \underline{0.5970} over the SOTA model, UNO. \textit{These results validate the effectiveness of ST-PAD as a purely data-driven approach in fluid modeling.}

\textbf{Obs.\ding{183} ST-PAD showcases stronger robustness and versatility than PINNs.} Since PINNs require knowledge of explicit and complete PDEs, in our experiments with real-world and video data, we are unable to determine the specific evolution parameters, thus preventing computation within PINNs. However, the design of ST-PAD only necessitates knowledge of general or partial physical equations at the upstream level. This underscores the robustness and versatility of ST-PAD (In our settings, we use conservation equations for KTH and TaxiBJ+, Wave Equation for SW, Navier-Stokes for NS and real-world meteorological datasets, Reaction-Convection-Diffusion Equation for CD benchmark.).

\textbf{Obs.\ding{184}} ST-PAD can effectively achieve higher prediction fidelity. From Fig \ref{fig:main1} and \ref{fig:main2}, it is clear that ST-PAD excels at capturing predictive details. Compared to traditional video prediction frameworks, such as PastNet and others, ST-PAD is adept at preserving the details in its predictions.

\begin{figure*}
  \centering
  \includegraphics[width=1\linewidth]{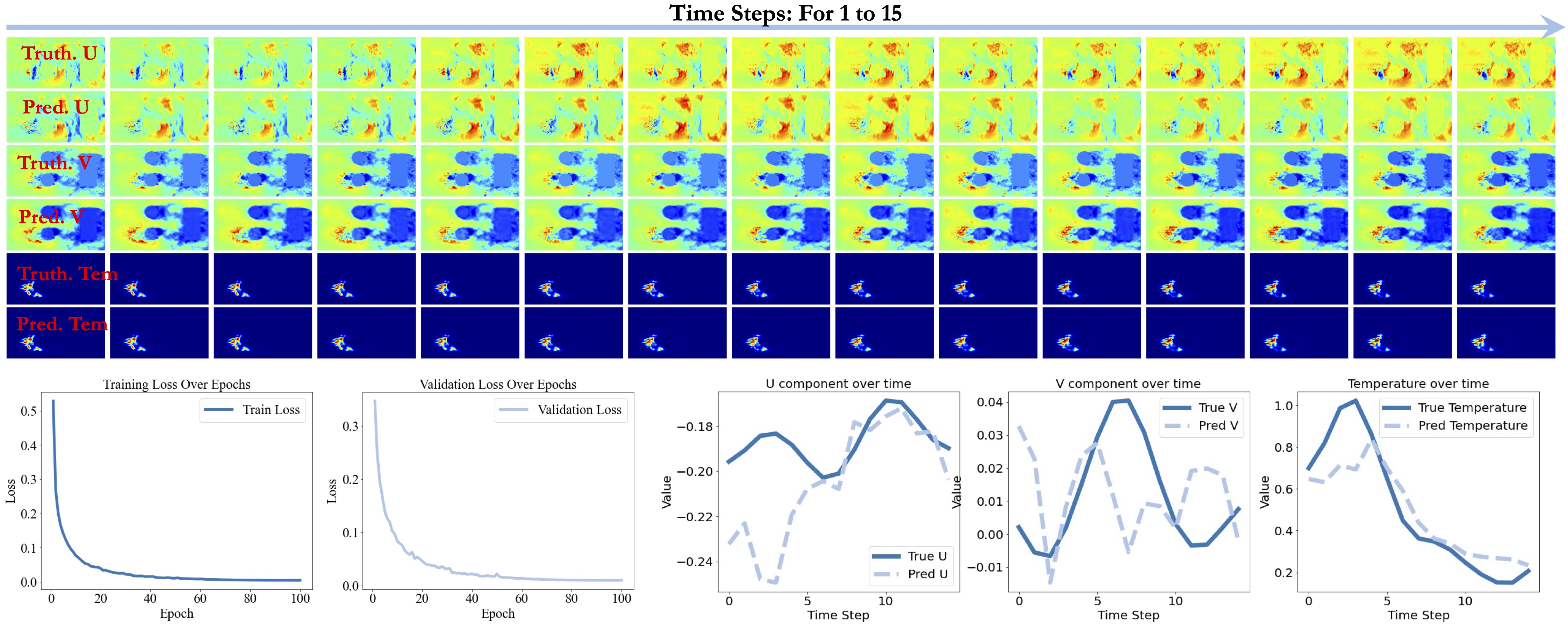}
       \caption{The figure shows the performance of the ST-PAD in simulating combustion dynamics data. The top three sets of images compare the true values (Truth) and predicted values (Pred) of U component, V component, and temperature over 15 time steps. The middle two graphs depict the decline in the loss function during training and validation across epochs. The bottom three graphs present the comparison between the true and predicted values of U component, V component, and temperature over time.}
  \label{fig:rq4-1}
\end{figure*}

\begin{figure}
  \centering
  \includegraphics[width=1\linewidth]{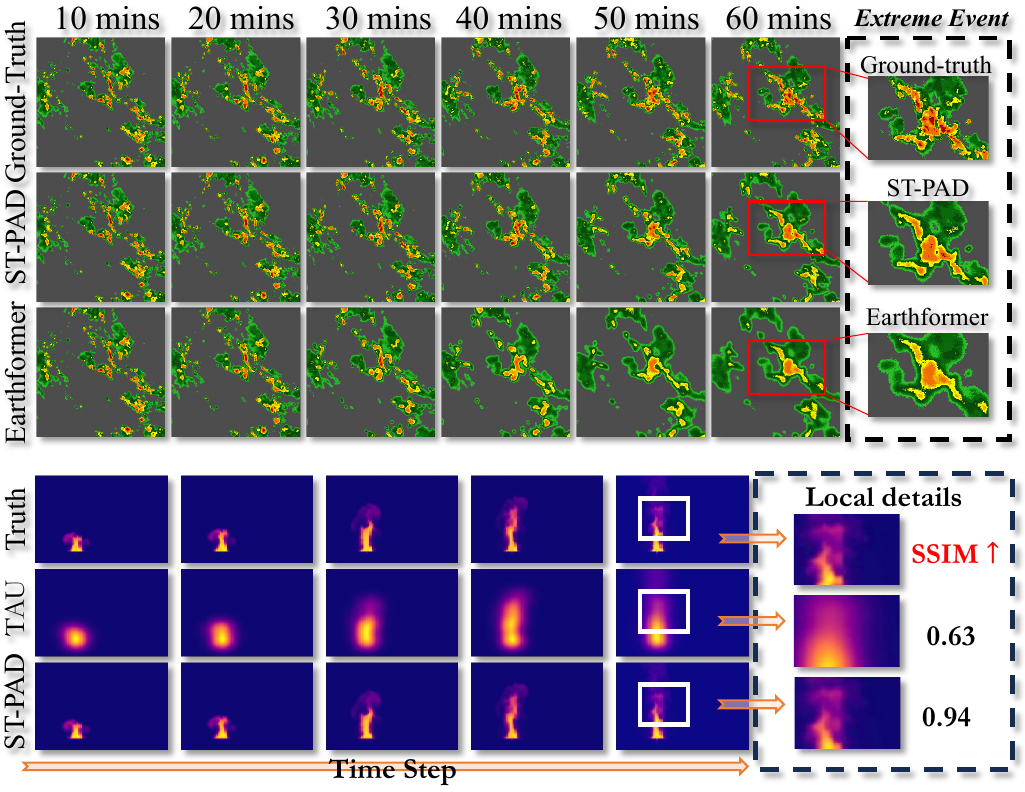}
       \caption{The top shows real and ST-PAD model predictions for 10 to 60-minute intervals, next to Earthformer model predictions. A red box highlights extreme events captured by both models and reality. Below are real temperature data and TAU and ST-PAD model predictions. An enlargement in the right-side white box shows details and SSIM scores, showing ST-PAD predictions closer to reality.}
  \label{fig:rq4-2}
\end{figure}

\subsection{Structured prediction results (RQ2)}
\vspace{-0.2em}

In this section, we further test whether ST-PAD can achieve more accurate structured prediction results compared to mainstream models. We employ PSNR and SSIM as prediction metrics and select MAD, Earthfarseer, FNO, and ST-PAD for comparison. We adopt two prediction approaches: autoregressive and parallel \cite{tan2023revisiting}. The autoregressive approach refers to recursive prediction, while parallel denotes simultaneous prediction. Employing these methods allows for a more comprehensive assessment of model stability across various scenarios. As shown in Tab \ref{tab:structure}, we can easily list the observations:

\textbf{Obs.\ding{185}} ST-PAD outperforms both PINNs and data-driven models on the TaxiBJ+, CD, and Weatherbench datasets. Notably, it achieves a PSNR increase of $1.33 \sim 5.25$ on the TaxiBJ+ dataset and a PSNR improvement of $2.54 \sim 3.04$ on the CD dataset. More importantly, in terms of the SSIM metric, ST-PAD perfectly restores the predictive effect, achieving excellent performance with scores over 0.95 on all three datasets. These results repeatedly validate ST-PAD's capability for structural prediction restoration and demonstrate its robustness.

\begin{table}[h]\footnotesize
  \centering
  \renewcommand{\arraystretch}{0.80}
    \setlength{\tabcolsep}{8.0pt}
  \caption{The PSNR and SSIM across different benchmarks and models.}
   \vspace{-0.5em}
    \begin{tabular}{c|c|cc}
    \toprule
    & \multirow{2}{*}{Models} & \multicolumn{2}{c}{PSNR/SSIM} \\
    & & Autoregressive & Parallel  \\
    \midrule
    \tabincell{c}{TaxiBJ+} & \tabincell{c}{ MAD \\ Earthfarseer \\ FNO \\ ST-PAD} & \tabincell{c}{ $23.58/0.68$\\ $38.94/0.96$ \\ $28.33/0.72$ \\ \underline{$40.27/0.98$}} & \tabincell{c}{$21.32/0.67$\\ $34.94/0.93$ \\ $24.46/0.69$ \\ \underline{$40.19/0.98$}}  \\
    \midrule
    CD  & \tabincell{c}{ MAD \\ Earthfarseer \\ FNO \\ ST-PAD} & \tabincell{c}{$26.87/0.77$\\ $40.21/0.92$ \\ $41.83/0.95$ \\ \underline{$44.37/0.99$}} & \tabincell{c}{$26.37/0.76$\\ $40.14/0.91$ \\ $40.42/0.94$ \\ \underline{$43.46/0.98$}} \\
    \midrule
    \tabincell{c}{Weatherbench} & \tabincell{c}{ MAD \\ Earthfarseer \\ FNO \\ ST-PAD} & \tabincell{c}{$17.34/0.54$\\ $37.43/0.93$ \\ $18.24/0.55$ \\ \underline{$38.34/0.95$}} & \tabincell{c}{$16.34/0.51$\\ $33.44/0.91$ \\ $19.23/0.57$ \\ \underline{$38.47/0.96$}}  \\

    \bottomrule
  \end{tabular}
  \label{tab:structure}%
\end{table}%

\subsection{Ablation Study (RQ3)}
 We conduct extensive ablation experiments to demonstrate the performance of the key components in \method{}, therefore, we propose the following model variants: (1) remove the upstream physical equation loss function (\method{} w/o $\textit{Physical}$); (2) remove the evolution module (\method{} w/o $\textit{Evolution}$); (3) remove the downstream parameter module (\method{} w/o $\textit{Parameter}$).  As shown in Table~\ref{tab:ablation}, The experimental results show that the complete \method{} model outperforms its variants on two benchmarks. Specifically, the full model reduces MAE to 2.0729 and increases SSIM to 0.9821 on the TaxiBJ+ benchmark, lowers MSE to 0.0875, and increases SSIM to 0.9643 on the Weatherbench benchmark. These findings highlight the crucial role of the physical equation loss function, evolution module, and parameter module in enhancing model performance.

\begin{table}[t]
    \caption{Ablation Studies on two dataset.}
    \vspace{-15pt}
    \small
    \label{tab:ablation}
    \vskip 0.15in
    \centering
    \begin{small}
            \renewcommand{\multirowsetup}{\centering}
            \setlength{\tabcolsep}{3.4pt} 
            \begin{tabular}{l|cc|cc}
                \toprule
                \multirow{3}{*}{Variants} & \multicolumn{4}{c}{Benchmarks}  \\
                \cmidrule(lr){2-5}
                & \multicolumn{2}{c}{TaxiBJ+} & \multicolumn{2}{c}{Weatherbench}   \\
                \cmidrule(lr){2-5}
               & MAE & SSIM & MSE & SSIM \\

                \midrule
                \method{} w/o $\textit{Physical}$  &2.3214 & 0.9521 & 0.0983 & 0.9312\\
                 \method{} w/o $\textit{Evolution}$ &2.2313 & 0.9623 & 0.0882 & 0.9533 \\
                \method{} w/o $\textit{Parameter}$ &2.4312 & 0.9321 & 0.0893 & 0.9442 \\
                \midrule
               \method{}  &2.0729 & 0.9821 & 0.0875 & 0.9643\\
                \bottomrule
            \end{tabular}
	\end{small}
    \vspace{-10pt}
\end{table}

\subsection{Local fidelity (RQ4)}

In the final subsection, we examine the local fidelity capabilities of ST-PAD, as local fidelity is crucial for fluid dynamics modeling. In Figure \ref{fig:rq4-1}, we observe the performance of the ST-PAD model in predicting the U and V components as well as temperature values compared to the true values over time steps $1 \sim 15$. We notice that the predicted curves of the U component closely align with the true U component for most time steps, despite some deviations at certain points, indicating that ST-PAD performs well in simulating the overall dynamics of the U component. Figure \ref{fig:rq4-2} highlights ST-PAD's exceptional performance in terms of local fidelity. Specifically, in predicting local extreme events, the ST-PAD model demonstrates superior capturing ability. For instance, at the 60-minute mark, the area of extreme events predicted by ST-PAD nearly perfectly matches the ground truth, while another compared model (Earthformer) underpredicts in the same area. In terms of TAU predictions, the comparison of temperature predictions by both models with the true data (Truth) shows that ST-PAD also excels in capturing local temperature peaks, with the SSIM score improving from 0.63-0.94. This indicates ST-PAD's significant advantage in preserving local details.

\section{Conclusion \& Future work}\label{sec:5}

In this section, we develop a two-stage framework named \textbf{ST-PAD}, which integrates spatio-temporal physical awareness and parameter diffusion guidance for continuum field modeling. Through our design, we introduce a Vector Quantized reconstruction module with time-evolution characteristics in the upstream phase, ensuring a balanced parameter distribution through universal physical constraints. In the downstream phase, we leverage a parameter-involved diffusion probabilistic network to generate high-quality future states of continuum fields while achieving parametric awareness of multiple physical settings. We conduct extensive experiments across multiple benchmarks covering different dynamical systems, and the experimental results demonstrate the effectiveness and robustness of our framework in capturing complex spatio-temporal dynamics and improving out-of-distribution generalization.

Modeling complex dynamical processes is important for understanding and addressing challenges in governance scenarios, particularly in areas such as environmental monitoring, traffic management, and disaster prevention. In this paper, we systematically investigate parameter-involved diffusion to enhance the generalization capability of continuum field modeling under diverse physical settings. The proposed framework provides a potential computational foundation for modeling dynamic processes and supporting data-driven governance applications. In the future, we plan to further incorporate generalization theories, such as invariant learning \cite{fang2023on,wang2023brave,eva,exgc,xia2024deciphering,zhou2023maintaining}, Mixture-of-Experts \cite{zhou2022mixture,du2022glam}, and adapter techniques \cite{wang2020k,chen2022vision}, to further improve generalization capabilities and explore their applicability to more complex governance scenarios.

\section{Acknowledgment}
This paper is partially supported by the National Natural Science Foundation of China (No.62072427, No.12227901), the Project of Stable Support for Youth Team in Basic Research Field, CAS (No.YSBR-005), Academic Leaders Cultivation Program, USTC.

\bibliographystyle{IEEEtran}
{
\bibliography{reference}
}





\ifCLASSOPTIONcaptionsoff
  \newpage
\fi

\end{document}